\documentclass{article}

\usepackage[final]{neurips_2019}

\usepackage[utf8]{inputenc} 
\usepackage[T1]{fontenc}    
\usepackage{url}            
\usepackage{booktabs}       
\usepackage{amsfonts}       
\usepackage{nicefrac}       
\usepackage{microtype}      
\usepackage{amsthm}
\usepackage{amsmath}
\usepackage{graphicx}
\usepackage{hyperref}       

\usepackage{amssymb}
\usepackage{amsmath}
\usepackage{enumerate}
\usepackage{enumitem}
\usepackage{xcolor}
\usepackage{comment}
\usepackage{float}
\usepackage[colorinlistoftodos]{todonotes}
\usepackage{times}  
\usepackage{helvet} 
\usepackage{courier}  
\usepackage{graphicx} 
\usepackage{graphicx}  
\title{Robust Variational Autoencoder}

\author{
   Haleh Akrami \\
   Signal and Image Processing Institute,\\ 
   Ming Hsieh Department of Electrical and Computer Engineering,\\ 
   University of Southern California,\\
   Los Angeles, CA, USA\\
   \texttt{email: akrami@usc.edu} \\
  \And
   Anand A. Joshi \\
   Signal and Image Processing Institute,\\ 
   Ming Hsieh Department of Electrical and Computer Engineering,\\ 
   University of Southern California,\\
   Los Angeles, CA, USA\\
   \texttt{email: ajoshi@usc.edu}
  \And
   Jian Li \\
   Signal and Image Processing Institute,\\ 
   Ming Hsieh Department of Electrical and Computer Engineering,\\ 
   University of Southern California,\\
   Los Angeles, CA, USA\\
   \texttt{email: jli981@usc.edu}
  \And
   Sergul Aydore \\
   Electrical and Computer Engineering,\\
   Stevens Institute of Technology,\\
   Hoboken, NJ, USA\\
   \texttt{email: sergulaydore@gmail.com}
  \And
   Richard M. Leahy \\
   Signal and Image Processing Institute,\\ 
   Ming Hsieh Department of Electrical and Computer Engineering,\\ 
   University of Southern California,\\
   Los Angeles, CA, USA\\
   \texttt{email: leahy@sipi.usc.edu}
}

\renewcommand{\vec}[1]{\mathbf{#1}}
\DeclareMathOperator{\E}{\mathbb{E}}
\begin{document}

\maketitle

\begin{abstract}
Machine learning methods often need a large amount of labeled training data. Since the training data is assumed to be the ground truth, outliers can severely degrade learned representations and performance of trained models. Here we apply concepts from robust statistics to derive a novel variational autoencoder that is robust to outliers in the training data. Variational autoencoders (VAEs) extract a lower-dimensional encoded feature representation from which we can generate new data samples. Robustness of autoencoders to outliers is critical for generating a reliable representation of particular data types in the encoded space when using corrupted training data. Our robust VAE is based on beta-divergence rather than the standard Kullback-Leibler (KL) divergence. Our proposed lower bound lead to a RVAE model that has the same computational complexity as the VAE and contains a single tuning parameter to control the degree of robustness. We demonstrate the performance of our $\beta$-divergence based autoencoder for a range of image datasets, showing improved robustness to outliers both qualitatively and quantitatively. We also illustrate the use of our robust VAE for outlier detection.
\end{abstract}

\section{Introduction}
Deep learning models that are based on the maximization of the log-likelihood, such as autoencoders \cite{hinton2006reducing}, assume training data as the ground truth. Outliers in training data can have a disproportionate impact on learning because they will have large negative log-likelihood values for a correctly trained network  \cite{gather1988maximum,huber2011robust}. In practice, and particularly in large datasets, training data will inevitably include mislabeled data, anomalies or outliers, sometimes taking up as much as  $10\%$ of the data \cite{hampel1986robust}. 

In the case of autoencoders, the inclusion of outliers in the training data can result in encoding of these outliers. As a result, the trained network may reconstruct these outliers in the testing samples. Conversely, if an encoder is robust to outliers then the outliers will not be reconstructed. Therefore, a robust autoencoder can be used to detect anomalies by comparing a test image to its reconstruction. 
 Here we focus on variational autoencoders (VAEs) \cite{kingma2013auto}. A VAE is a probabilistic graphical model that is comprised of an encoder and a decoder. The encoder transforms high-dimensional input data with an intractable probability distribution into a low-dimensional `code' with an approximate posterior (variational distribution) that is tractable. The decoder then samples from the posterior distribution of the code and transforms that sample into a reconstruction of the input. VAEs use the concept of variational inference \cite{bishop2006pattern} and re-parameterize the variational evidence lower bound (ELBO) so that it can be optimized using standard stochastic gradient descent methods.

A VAE can learn latent features that best describe the distribution of the data and allows the generation of new samples using the decoder. VAEs have been successfully used for feature extraction from images, audio and text \cite{kusner2017grammar,pu2016variational,hsu2017learning}. Moreover, when VAEs are trained using normal datasets, they can be used to detect anomalies, where the characteristics of the anomalies differ from those of the training data \cite{an2015variational,pmlr-v102-you19a}. 
 
\subsection{Related Work} 
In the past few years, denoising autoencoders \cite{vincent2008extracting}, maximum correntropy autoencoders \cite{qi2014robust} and robust autoencoders \cite{zhou2017anomaly} have been proposed to overcome the problem of noise corruption, anomalies, and outliers in the data. The denoising autoencoder \cite{vincent2008extracting} is trained to reconstruct `noise-free' inputs from corrupted data and is robust to the type of corruption it learns. However, these denoising autoencoders require access to clean training data and the modeling of noise can be difficult in real-world problems. An alternative approach is to replace the cost function with noise-resistant correntropy \cite{qi2014robust}. Although this approach discourages the reconstruction of outliers in the output, it may not prevent encoding of outliers in the hidden layer. Recently, \cite{zhou2017anomaly} described a robust deep autoencoder that was inspired by robust principal component analysis. This encoder performs a decomposition of input data $X$ into two components, $X = LD + S$, where $LD$ is the low-rank component which we want to reconstruct and $S$ represents a sparse component that contains outliers or noise.

Despite many successful applications of these models, they are not probabilistic and hence do not extend well to generative models. Generative models learn distributions from the training data allowing generation of novel samples that match the training samples' characteristics.

\subsection{Motivation via a Toy Example} 
\begin{figure}[ht]
    \centering
    \includegraphics[width=.9\linewidth]{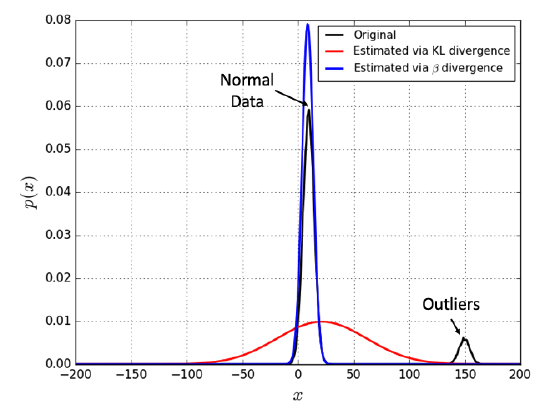}
    \caption{Illustration of robustness of $\beta$-divergence to outliers: optimizing KL-divergence for parameter estimation of Gaussian attempts to also account for the presence of outliers, whereas optimizing $\beta$-divergence results in an estimate that is more robust to outliers.}
    \label{fig:illustration}
\end{figure}

In order to make the VAE models robust to the outliers in the training data, existing approaches focus on modification of network architectures, adding constraints or modeling of outlier distribution \cite{zhai2017robust,cao2019coupled,eduardo2019robust}. In contrast, we adopt an approach based on the $\beta$-divergence from robust statistics \cite{basu1998robust}. To motivate our approach on using $\beta$-divergence, we ran simple simulations (Figure \ref{fig:illustration}). Here, the samples were generated from a distribution $p$ which is a mixture of two Gaussian distributions where the tall mode represents normal samples and the short mode indicates the presence of outliers. Our goal was to learn a single-mode Gaussian distribution $q_{\theta}$ by minimizing KL and $\beta$ divergences separately to optimize $\theta$. Figure \ref{fig:illustration} shows the estimated distributions optimized for the two different divergences. 
The $\beta$-divergence estimate is robust to the outliers whereas the estimated distribution from the KL divergence attempts to also account for the presence of outliers. 

\subsection{Our Contributions} We propose a novel robust VAE (RVAE) using robust variational inference \cite{futami2017variational} that uses a $\beta-$ELBO based cost function. The $\beta$-ELBO cost replaces the  KL-divergence (log-likelihood) term  with $\beta$-divergence. Our contributions:
\begin{itemize}
    \item We apply concepts from robust statistics, specifically, robust variational inference to variational autoencoder (VAE) for deriving a robust variational autoencoder (RVAE) model. We also present  formulations of RVAE for both Gaussian and Bernoulli cases.
    \item We show that on benchmark datasets from computer vision and real-world brain imaging datasets our approach is more robust than a standard VAE to outliers in the training data.
    \item We also show how the robustness of RVAE can be exploited to perform anomaly detection even in cases where the training data also includes similar anomalies.
\end{itemize}

\section{Mathematical Formulation}

Let $\mathbf{x}^{(i)} \in \mathbb{R}^D$ be an observed sample of input $\mathbf{X}$ where $i \in \{1, \cdots, N \}$, $D$ is the number of features and $N$ is the number of samples; and $\mathbf{z}^{(j)}$ be an observed sample for latent variable $\mathbf{Z}$ where $j \in \{1, \cdots, S \}$.
Given samples $\vec{x}^{(i)}$ of the random feature vector  $\vec{X}$ representing input data, probabilistic graphical models estimate the posterior distribution $p_{\theta}(\vec{Z}|\vec{X})$ as well as the model evidence $p_{\theta}(\vec{X})$, where $\vec{Z}$ represents the latent variables and $\theta$ the generative model parameters \cite{bishop2006pattern}. The goal of variational inference is to approximate the posterior distribution of $\vec{Z}$ given $\vec{X}$ by a tractable parametric distribution. In variational methods, the functions used as prior and posterior distributions are restricted to those that lead to tractable solutions. For any choice of a tractable distribution $q(\vec{Z})$, the distribution of the latent variable, the following decomposition holds:
\begin{align}
\log p_{\theta}(\vec{X})=L(q(\vec{Z}),{\theta})+ D_{KL}(q(\vec{Z})||p_{\theta}(\vec{Z}|\vec{X})),
\end{align}
where $D_{KL}$ represents the Kullback-Leibler (KL) divergence. Instead of maximizing the log-likelihood $p_{\theta}(\vec{X})$, with respect to the model parameters $\theta$, the variational inference approach maximizes its variational evidence lower bound ELBO \cite{bishop2006pattern}:
\begin{align*}
L(q,{\theta}) &= \E_{q(\vec{Z})}[\log(p_{\theta}(\vec{X}|\vec{Z}))]-D_{KL}(q(\vec{Z})||p_{\theta}(\vec{Z})).
\end{align*}

\subsection{Robust Variational Inference}
\label{sec:robust_vi}
Here we review the robust variational inference framework \cite{futami2017variational} and explain its usage for developing variational autoencoders that are robust to outliers. The ELBO function includes a log-likelihood term which is sensitive to outliers in the data because the negative log-likelihood of low probability samples can be arbitrarily high.
It can be shown that maximizing log-likelihood given samples $\mathbf{x}^{ (i) }$ is equivalent to minimizing KL divergence $D_{KL}(\hat{p}(\vec{X})||p_{\theta}(\vec{X}|\vec{Z}))$
between the empirical distribution $\hat{p}$ of the samples and the parametric
distribution $p_{\theta}$ \cite{zellner1988optimal,futami2017variational}.
Therefore, the ELBO function can be expressed as:
\begin{align}
\begin{split}
L(q,{\theta})=&-N\E_{q}\left[D_{KL}(\hat{p}(\vec{X})||p_{\theta}(\vec{X}|\vec{Z}))\right]\\&-D_{KL}(q(\vec{Z})||p_{\theta}(\vec{Z})) + \mbox{const.},
\end{split}
\end{align}
where $N$ is the number of samples of $\vec{X}$ used for computing the empirical distribution $\hat{p}(\vec{X}) = \frac{1}{N} \sum_{i=1}^N \delta(\vec{X}, \mathbf{x}^{ (i) })$ and $\delta$ is the Dirac delta function. Rather than using KL divergence, which is not robust to outliers, it is possible to choose a different
divergence measure to quantify the distance between two distributions. Here we use $\beta$-divergence, $D_{\beta}$ \cite{basu1998robust}: 
\begin{align*}
\begin{split}
& D_{\beta}(\hat{p}(\vec{X})||p_{\theta}(\vec{X}|\vec{Z}))= \frac{1}{\beta}\int_{X}\hat{p}(\vec{X})^{\beta+1} d\vec{X}\\ & - \frac{\beta+1}{\beta}\int_{\vec{X}}\hat{p}(\vec{X})p_{\theta}(\vec{X}|\vec{Z})^{\beta} d\vec{X} + \int_{\vec{X}}p_{\theta}(\vec{X}|\vec{Z})^{\beta+1} d\vec{X}.
\end{split}
\end{align*} 
In the limit as ${\beta\to 0}$, $D_{\beta}$ converges to $D_{KL}$. Using $\beta$-divergence changes the variational inference   optimization problem to maximizing $\beta$-ELBO:
\begin{align}
\begin{split}
L_{\beta}(q,{\theta})=&-N \E_{q}[(D_{\beta}(\hat{p}(\vec{X})||p_{\theta}(\vec{X}|\vec{Z})))]\\&-D_{KL}(q(\vec{Z})||p_{\theta}(\vec{Z}))
\end{split}
\label{eq:beta-elbo}
\end{align}
Note that for robustness to the outliers in the input data, the divergence in the likelihood is replaced, but divergence in the latent space is still the same \cite{futami2017variational}. The idea behind $\beta$-divergence is based on applying a power transform to variables with heavy tailed distributions \cite{cichocki2010families}.  It can be proven that minimizing $D_{\beta}(\hat{p}(\vec{X})||p_{\theta}(\vec{X}|\vec{Z}))$ is equivalent to minimizing  $\beta$-cross entropy \cite{futami2017variational}, which, in the limit of ${\beta\to 0}$, converges to the cross entropy, and is given by \cite{eguchi2010entropy}:
\begin{align}
\begin{split}
& H_{\beta}(\hat{p}(\vec{X}),p_{\theta}(\vec{X}|\vec{Z}))  =  \\ &- \frac{\beta+1}{\beta}  \int \hat{p}(\vec{X})({p_{\theta}(\vec{X}|\vec{Z})^{\beta}-1})d\vec{X}  + \int{p_{\theta}(\vec{X}|\vec{Z})^{\beta+1}}d\vec{X}.
\end{split}
\label{eq:cross_entropy_empirical}
\end{align}
Replacing $D_{\beta}$ in eq. \ref{eq:beta-elbo} with $H_{\beta}$ results in
\begin{align}
\begin{split}
L_{\beta}(q,{\theta}) = & -N \E_{q}[(H_{\beta}(\hat{p}(\vec{X})||p_{\theta}(\vec{X}|\vec{Z})))]\\ & - D_{KL}(q(\vec{Z})||p_{\theta}(\vec{Z})).
\end{split}
\label{eq:beta-elbo2}
\end{align}

\subsection{Variational Autoencoder}
\label{sec:vae}
A variational autoencoder (VAE) is a directed probabilistic graphical model whose posteriors are approximated by a neural network. It has two components: the encoder network that computes $q_{\phi}(\vec{Z}|\vec{X})$, which is a tractable approximation of the intractable posterior $p_{\theta}(\vec{Z}|\vec{X})$, 
and the decoder network that computes $p_{\theta}(\vec{X}|\vec{Z})$, which together form an autoencoder-like architecture \cite{wingate2013automated}. 
The regularizing assumption on the latent variables is that the marginal $p_\theta(\vec{Z})$ is a standard Gaussian $N(0,1)$. For this model the marginal likelihood of individual data points can be rewritten as follows:
\begin{align}
\begin{split}
\log p_{\theta}(\vec{x}^{(i)})  = & D_{KL}(q_{\phi}(\vec{Z}|\vec{x}^{(i)}),p_{\theta}(\vec{Z}|\vec{x}^{(i)}))\\
 & + L(\theta,\phi;\vec{x}^{(i)}),
\end{split}
\end{align}
where
\begin{align}
\begin{split}
    L(\theta,\phi;\vec{x}^{(i)})=& \E_{q_{\phi}(\vec{Z}|\vec{x}^{(i)})}[\log(p_{\theta}(\vec{x}^{(i)}|\vec{Z}))]\\ &-D_{KL}(q_{\phi}(\vec{Z}|\vec{x}^{(i)})||p_{\theta}(\vec{Z})).
\end{split}
\end{align}

The first term (log-likelihood) can be interpreted as the  \emph{reconstruction loss} and the
second term (KL divergence) as the \emph{regularizer}. Using empirical estimates of expectation we form the Stochastic Gradient Variational Bayes (SGVB) cost \cite{kingma2013auto}:
\begin{align}
\begin{split}
L(\theta,\phi;\vec{x}^{(i)})\approx &\frac{1}{S}\sum_{j=1}^{S}\log(p_{\theta}(\vec{x}^{(i)}|\vec{z}^{(j)}))\\ & -D_{KL}(q_{\phi}(\vec{Z}|\vec{x}^{(i)})||p_{\theta}(\vec{Z})), \label{eq:vae_elbo}
\end{split}
\end{align}
where $S$ is the number of samples drawn from $q_{\phi}(\vec{Z}|\vec{X})$.
We can assume either a multivariate i.i.d. Gaussian or Bernoulli distribution for $p_{\theta}(\vec{X}|\vec{Z})$. That is, given the latent variables, the uncertainty remaining in $\vec{X}$ is i.i.d. with these distributions. For the Bernoulli case,  the log likelihood for sample $\vec{x}^{(i)}$ simplifies to:
\begin{align*}
\begin{split}
& \E_{q_{\phi}}(\log p_{\theta}(\vec{x}^{(i)}|\vec{Z}))\approx  \frac{1}{S}\sum_{j=1}^{S}\log p_{\theta}(\vec{x}^{(i)}|\vec{z}^{(j)})\\ & =\frac{1}{S}  \sum_{j=1}^{S}\sum_{d=1}^{D}\left(\vec{x}^{(i)}\log p_{d}^{(j)}+(1-\vec{x}_d^{(i)})\log(1-p_{d}^{(j)})\right),
\end{split}
\end{align*}
where $p_\theta(\vec{x}_d^{(i)}|\vec{z}^{(j)}) = \text{Bernoulli}(p^{(j)}_d)$ and $D$ is the feature dimension. In practice we can choose $S=1$ as long as the minibatch size is large enough. For the Gaussian case, this term simplifies to the mean-squared-error.


\subsection{Robust Variational Autoencoder}
We now derive the robust VAE (RVAE) using concepts discussed above. In order to derive the cost function for the RVAE, as in eq. \ref{eq:beta-elbo2}, we propose to use $\beta$-cross entropy $H_{\beta}^{(i)}(\hat{p}(\vec{X}), p_{\theta}(\vec{X}|\vec{Z}))$ between the empirical distribution of the data $\hat{p}(\vec{X})$ and the probability of the samples for the generative process $p_{\theta}(\vec{X}|\vec{Z})$ for each sample $\vec{x}^{(i)}$ in place of the likelihood term in eq. \ref{eq:vae_elbo} . Similar to VAE, the regularizing assumption on the latent variables is that the marginal $p_\theta(\vec{Z})$ is normal Gaussian $N(0,1)$. the $\beta$-ELBO for the RVAE is: 
\begin{align*}
\begin{split}
L_{\beta}(\theta,\phi;\vec{x}^{(i)}) = & 
-\E_{q_{\phi}(\vec{Z}|\vec{x^{(i)}})}[(H_{\beta}^{(i)}(\hat{p}(\vec{X}), p_{\theta}(\vec{X}|\vec{Z})))]\\ & -D_{KL}(q_{\phi}(\vec{Z}|\vec{x}^{(i)})||p_{\theta}(\vec{Z})).
\end{split}
\end{align*}
\begin{figure*}
    \centering
    \includegraphics[width=.95\linewidth]{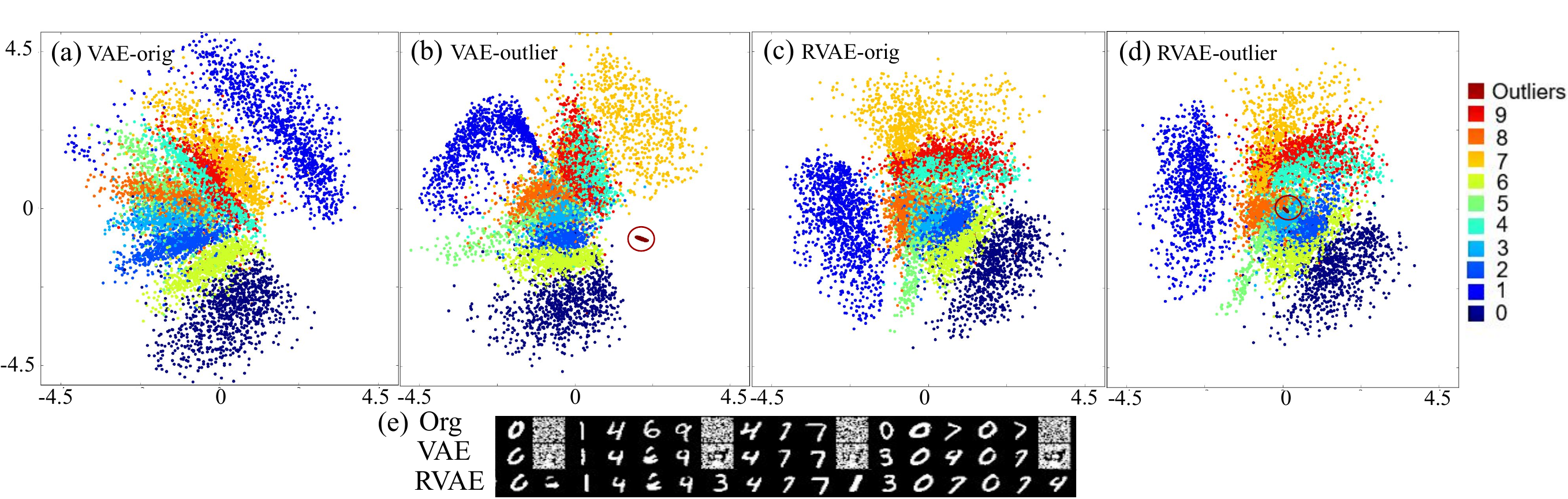}
    \caption{Comparing robustness of VAE and our RVAE using the MNIST dataset contaminated with synthetic outliers generated by Gaussian noise:  (a) the 2D latent space of VAE for the original MNIST dataset without outliers (colors represent class labels of MNIST); (b) the 2D latent space of VAE for the MNIST dataset with added outliers (marked by a dark red circle); (c) the 2D latent space of RVAE without outliers; (d) the 2D latent space of RVAE with outliers added to the input data. (e) Examples of the reconstructed images using VAE and RVAE. Unlike the VAE, our RVAE is minimally affected by the presence of outliers and did not reconstruct the outliers (noise images).}
    \label{fig:2D}
\end{figure*}
\subsubsection{Bernoulli Case}
For each sample we need to calculate $H_{\beta}^{(i)}(\hat{p}(\vec{X}) , p_{\theta}(\vec{X}|\vec{Z}))$ when $\vec{x}^{(i)}\in\{0,1\}$. Using empirical estimates of expectation we
form the SGVB and chose S=1. In eq. \ref{eq:cross_entropy_empirical} we substitute  $\hat{p}(\vec{X})=\delta(\vec{X}-\vec{x}^{(i)})$  and $p_{\theta}(\vec{X}|\vec{z}^{(j)})$ is a Bernoulli distribution therefore $p_{\theta}(\vec{X}|\vec{z}^{(j)})^\beta=\left(\vec{X}p^{(j)}+(1-\vec{X})(1 - p^{(j)} )\right)^{\beta} = \vec{X}p^{(j)\beta}+(1-\vec{X})\left(1-p^{(j)}\right)^{\beta}$, and 
{\small
\begin{align*}
\begin{split}
&H_{\beta}^{(i)} (\hat{p}(\vec{X}), p_{\theta}(\vec{X}|\vec{z}^{(j)}))=\\& -\frac{\beta+1}{\beta}
\int \delta(\vec{X}-\vec{x}^{(i)}) (\vec{X}p^{(j)}+(1-\vec{X})(1-p^{(j)})^{\beta}-1) d\vec{X} \\
 &+ p^{(j)\beta+1}+{(1-p^{(j)})}^{\beta+1}. 
\end{split}
\end{align*}
}
For the second integral, we calculate the sum over $\vec{x}^{(i)}\in\{0,1\}$. Therefore, for the multivariate
case, the $\beta$-ELBO of RVAE becomes:
{\small
\begin{align}
\begin{split}
& L_{\beta}(\theta,\phi;\vec{x}^{(i)}) = \\ & \frac{\beta+1}{\beta}\left(\prod_{d=1}^{D}\left(\vec{x}_{d}^{(i)}p_{d}^{(j)\beta}+(1-\vec{x}_{d}^{(i)})(1-p_{d}^{(j)})^{\beta}\right)-1\right)\\
- & \prod_{d=1}^{D}\left(p_{d}^{(j)\beta+1}+(1-p_{d}^{(j)\beta+1})\right) -D_{KL}(q_{\phi}(\vec{Z}|\vec{x}^{(i)})||p_{\theta}(\vec{Z})).
\end{split}
\label{eq:bernoulli_ELBO}
\end{align}
}
The Bernoulli assumption is useful when the data is binary. 

\begin{figure*}[ht]
    \centering
    \includegraphics[width=1\linewidth]{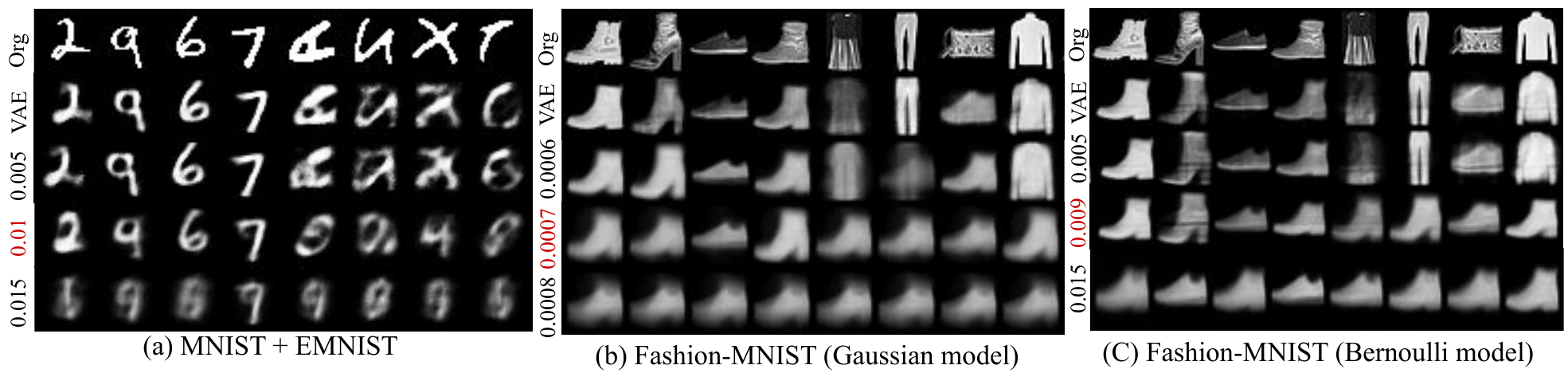}

    \caption{Examples of reconstructed normal images (first 4 columns in each Figure) and outliers (last 4 columns in each Figure) using VAE and RVAE with different $\beta$s on (a) MNIST (normal) + EMNIST (outliers) datasets and (b,c) Images from the class of shoes (normal) and images from the class of other accessories (outliers) in the Fashion MNIST datasets. The optimal value of $\beta$ is highlighted in \textcolor{red}{red}. The RVAE with too small $\beta$ has similar performance to the regular VAE, while RVAE with too large $\beta$ rejects outliers but also rejects some normal samples.}
    \label{fig:recons}
\end{figure*}

\subsubsection{Gaussian Case}
When the data is continuous and unbounded we can assume the posterior $p(\vec{X}|\vec{z}^{(j)})$ is Gaussian  $N(\hat{\vec{x}}^{(j)},\sigma)$ where $\hat{\vec{x}}^{(j)} $ is the output of the decoder generated from $\vec{z}^{(j)}$. Here, we choose $\sigma=1$ for our experiments. While we empirically found that the value of $\sigma$ does not have a significant impact on the performance of the autoencoder, the choice of $\sigma$ needs to be analyzed further. The $\beta$-cross entropy for the $i^{th}$ sample input $\vec{x}^{(i)}$ is given by:
\begin{align}
\begin{split}
& H_{\beta}^{(i)}(\hat{p}(\vec{X}), p_{\theta}(\vec{X}|\vec{z}^{(j)})) = \\ &-\frac{\beta+1}{\beta}\int\delta(\vec{X}-\vec{x}^{(i)})(N(\hat{\vec{x}}^{(j)},\sigma)^{\beta}-1)d\vec{X}\\
&+\int N(\hat{\vec{x}}^{(j)},\sigma)^{\beta+1}d\vec{X}.
\end{split}
\end{align}
The second term does not depend on $\hat{\vec{x}}$ so the first term is
minimized when \mbox{$\exp(-\beta\sum_{d=1}^{D}||\hat{\vec{x}}_{d}^{(j)}-\vec{x}_{d}^{(i)}||^{2})$}
is maximized.
Therefore, the ELBO-cost for the Gaussian case for $j^{th}$ sample is then given by
{\small
\begin{align}
\begin{split}
&L_{\beta}(\theta,\phi;\vec{x}^{(i)})= \\& \quad \frac{\beta+1}{\beta} \left(\frac{1}{(2\pi\sigma^{2})^{\beta D/2}}\exp \left(-\frac{\beta}{2\sigma^{2}}\sum_{d=1}^{D}||\hat{\vec{x}}_{d}^{(j)}-\vec{x}_{d}^{(i)}||^{2} \right)-1 \right) \\
& \quad -D_{KL}(q_{\phi}(\vec{Z}|\vec{x}^{(i)})||p_{\theta}(\vec{Z})).
\end{split}
\label{eq:Gaussian_ELBO}
\end{align}
}
In the following sections, we used the Bernoulli formulation for Experiments 1 and 2 which used the MNIST, EMNIST, and the Gaussian formulation for  Fashion-MNIST dataset and brain imaging data in Experiment 3. In each case we optimized the cost function using stochastic gradient descent with reparameterization \cite{kingma2013auto}.

\section{Experimental Results}
Here we evaluate the performance of RVAE using datasets contaminated with outliers and compare with the traditional VAE. We conducted three experiments using the MNIST  \cite{lecun1998gradient}, the EMNIST  \cite{cohen2017emnist}, the Fashion-MNIST benchmark datasets \cite{xiao2017fashion}, and two real-world brain imaging datasets: Maryland MagNeTs study of neurotrauma (\url{https://fitbir.nih.gov}) and The Ischemic Stroke Lesion Segmentation (ISLES) database \cite{maier2017isles} (\url{http://www.isles-challenge.org}). Both brain imaging data datasets consist of three sets of coregistered images corresponding to FLAIR and T1 and T2 weightings. We used fully-connected deep neural networks for the encoder and the decoder in the first two experiments.
 The first two experiments consisted of an encoder and decoder that are both single fully-connected layers with 400 units and a hidden layer in between. We used a deeper architecture for Experiment 3 described in \cite{larsen2015autoencoding} to capture the higher complexity of the data. The size of latent dimensions in the bottleneck layer was chosen based on the size and complexity of the datasets.
We used PyTorch \cite{paszke2017automatic} for the implementation and our code is in the supplemental material. We used the Adam algorithm \cite{kingma2014adam} with a learning rate of 0.001 for training.
\begin{figure*}[ht]
    \centering
    \includegraphics[width=1\linewidth]{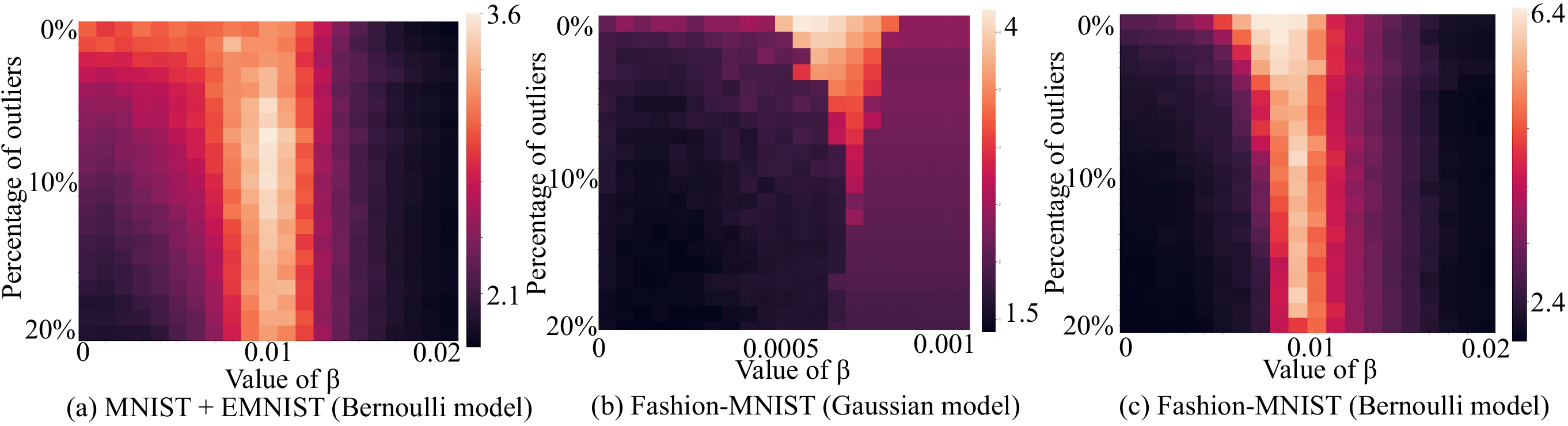}
    \caption{Heat maps of the performance measure as a function of the parameter $\beta$ (x-axis) and the fraction of outliers present in the training data (y-axis) for two datasets used for experiment 2.}
    \label{fig:heatmap_roc}
\end{figure*}
\begin{figure*}[ht]
    \centering
    \includegraphics[width=.95\linewidth]{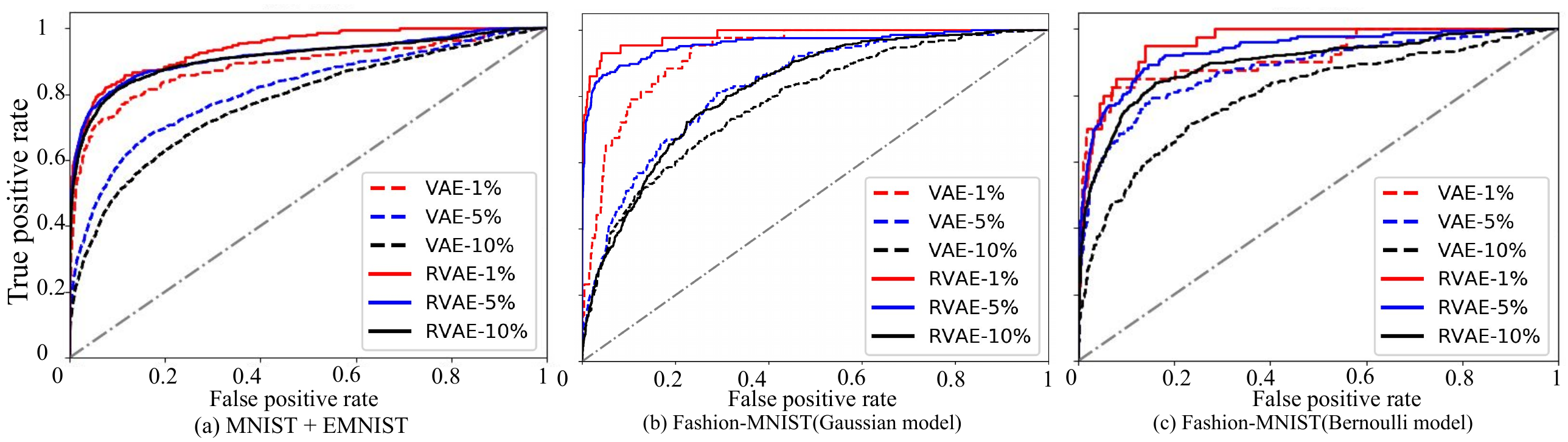}
    \caption{ROC curves showing the performance of outlier detection using VAE and RVAE with different fraction of outliers present in the training data for two datasets used in the experiment 2. The RVAE outperforms the VAE for all settings and the difference becomes larger as the fraction of outliers increases.}
    \label{fig:rocs}
\end{figure*}

\subsection{Experiment 1: Effect on Latent Representation}
First, we used the MNIST dataset comprising 70,000 28x28 grayscale images of handwritten digits \cite{lecun1998gradient}. We replaced $10\%$ of the MNIST data with synthetic outlier images generated by white Gaussian noise.  We binarized the data by thresholding at $0.5$ of the maximum intensity value, and used the Bernoulli model of the $\beta$-ELBO (eq.  \ref{eq:bernoulli_ELBO}) with $\beta=0.005$. The latent dimension was chosen to be $2$ for visual inspection. Figure~\ref{fig:2D} (e) shows examples of the reconstructed images using the VAE (second row) and RVAE (third row) along with original images (first row). The results show that outlier images are encoded when VAE is used. On the other hand, our RVAE did not encode the outlier noise images and encoded them to produce images consistent with the MNIST training data. Moreover, we visually inspected the embeddings using both VAE and RVAE (Figure~\ref{fig:2D} (a) - (d)). In the VAE case (Figure~\ref{fig:2D} (a) and (b)), the distributions of the digits were strongly perturbed by the outlier noise images. In contrast, RVAE was not significantly affected by outliers (Figure~\ref{fig:2D} (c) and (d)), illustrating the robustness of our RVAE.
\begin{figure*}[ht]
    \centering
    \includegraphics[width=1\linewidth]{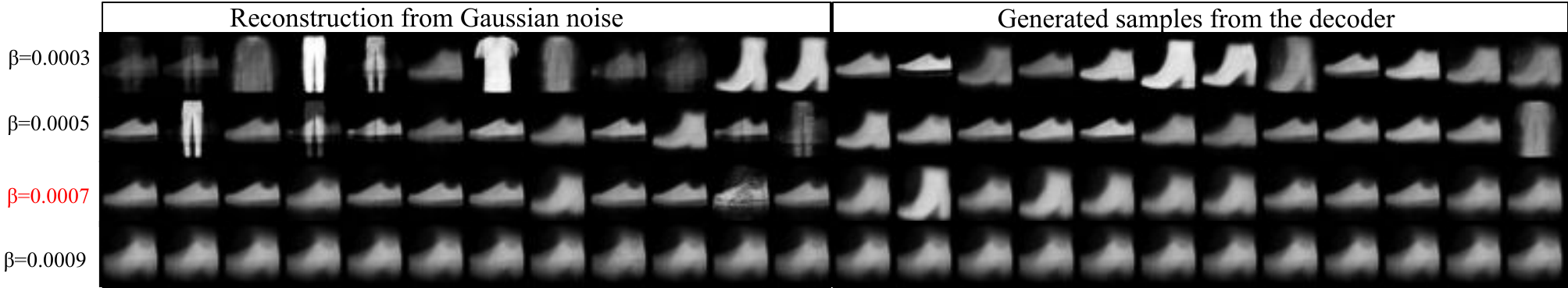}
    \caption{Searching for best $\beta$ by visual inspection: we compare the reconstructed images of the fake outliers generated by Gaussian noise with the generated samples. The optimal value of $\beta$ is $0.0007$ since the reconstructed images and the generated samples look more similar and there is a variability in generated samples. This optimal value also matches the maximum value achieved in the heat map in Figure \ref{fig:heatmap_roc}-b.}
    \label{fig:comparison_beta_fakeoutliers}
\end{figure*}
 \begin{figure*}
    \centering
    \includegraphics[width=1\linewidth]{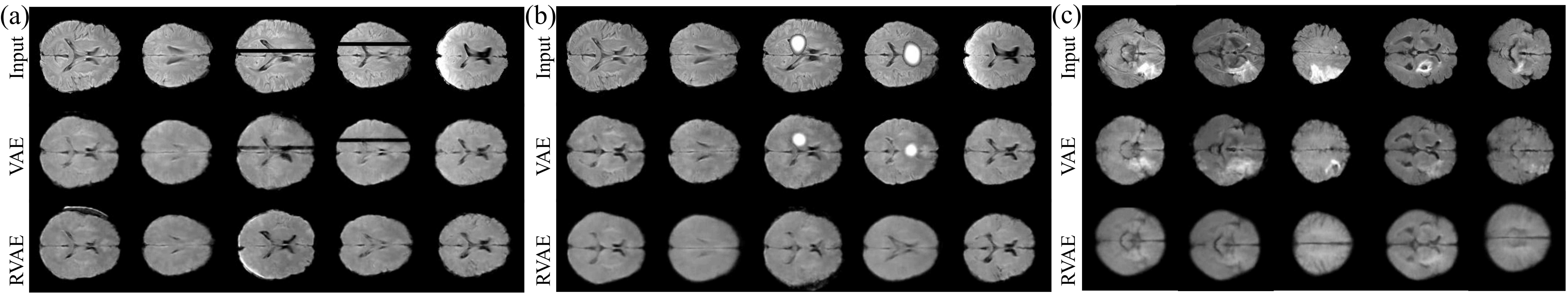}
    \caption{Reconstructions of brain images using VAE and RVAE: a) after randomly dropping  rows with a height of 5 pixels for 10\% of the Maryland MagNeTs dataset; b) after adding random lesion to 10 percent of the Maryland MagNeTs dataset; c) on the ISEL data with true lesions. Unlike VAE, both simulated and real outliers are not reconstructed by the RVAE.}
    \label{fig:brainlesions}
\end{figure*}
\subsection{Experiment 2: Reconstruction and Outlier Detection}
Second instead of using Gaussian random noise as outliers, we replaced a fraction of the MNIST data with Extended MNIST (EMNIST) data \cite{cohen2017emnist} which contains images that are the same size as MNIST. We binarized the data by thresholding at $0.5$ of the maximum intensity value, and used the Bernoulli model of the $\beta$-ELBO (eq.  \ref{eq:bernoulli_ELBO}) for the RVAE loss function. 

Similarly, we repeated the above experiment using the Fashion-MNIST dataset \cite{xiao2017fashion} that consists of 70,000 28x28 grayscale images of fashion products from 10 categories (7,000 images per category).  Here we chose shoes and sneakers as normal classes and samples from other categories as outliers. Since these images contain a significant range of gray scales, we chose the Gaussian model for the $\beta$-ELBO (eq.  \ref{eq:Gaussian_ELBO}). An apparently common, yet theoretically unclear practice is to use a Bernoulli model for grayscale data. This is pervasive in VAE tutorials, research literature and default implementations of VAE in deep learning frameworks where researchers, effectively treat the data as probability values rather than samples from the distribution . This issue is discussed in detail in \cite{loaiza2019continuous} with a possible solution based on a `continuous Bernoulli' distribution.  To be consistent with this practice, we also include results for the Fashion-MNIST grayscale images using the Bernoulli model of the $\beta$-ELBO (eq. \ref{eq:bernoulli_ELBO}) for our RVAE loss function.

 To investigate the performances of the autoencoders, we start with a fixed fraction of outliers (10\%). For MNIST-EMNIST experiment we trained both VAE and RVAE with $\beta$ varying from 0.001 to 0.02. Figure \ref{fig:recons} (a) shows the reconstructed images from RVAE with $\beta=0.005, 0.01$ and $0.015$ in comparison with the regular VAE. Similarly to experiment 1, with an appropriate $\beta$ ($\beta=0.01$ in this case), RVAE did not reconstruct the outliers (letters). As expected, RVAE with too small $\beta$ has similar performance to the regular VAE, while RVAE with too large $\beta$ rejects outliers but also rejects some normal samples.

Next, we explored the impact of the parameter $\beta$ and the fraction of outliers in the data on the performance of the RVAE. The performance was measured as the ratio between the overall absolute reconstruction error in outlier samples (letters) and their counterparts in the normal samples (digits). The higher this metric, the more robust the model, since a robust model should in this example encode digits well but letters (outliers) poorly. Figure~\ref{fig:heatmap_roc} (a) shows the performance of this measure in a heatmap as a function of $\beta$ (x-axis) and the fraction of outliers (y-axis). 
When only a few outliers are present, a wide range of  $\beta$s ($<0.01$) works almost equally well. On the other hand, when a significant fraction of the data is outliers, the best performance was achieved only when $\beta$ is close to $0.01$. When $\beta>0.01$, the performance degraded regardless of the fraction of the outliers. These results are consistent with the results in  Figure~\ref{fig:recons}.

We further investigate the performance of RVAE as a method for outlier detection. For this purpose, we threshold the mean squared error between the reconstructed images and the original images to detect outliers. The resulting labels were compared to the ground truth to determine true and false-positive rates.  We varied the threshold to compute Receiver Operating Characteristic (ROC) curves. Figure \ref{fig:rocs} shows the ROC curves with RVAE shown as a solid line and VAE shown as a dashed line. The results were similar for Fashion-MNIST dataset (Figures (\ref{fig:recons},\ref{fig:heatmap_roc},\ref{fig:rocs}) (b),(c)).  
The RVAE outperformed the VAE for all settings and the difference becomes larger as the fraction of outliers increases.

\subsection{How to choose robustness parameter $\beta$?}
In practice, outliers in the training data are not known in advance, hence we cannot compute the reconstruction error in outlier samples as described above. Instead, we propose to use \textit{fake} outliers generated by Gaussian noise and inspect the reconstructed images. At the same time, we visualize the generated samples from the decoder. In Figure \ref{fig:comparison_beta_fakeoutliers}, we show these two results for RVAE models trained with different $\beta$ values from a Gaussian model using the Fashion-MNIST dataset with $10 \%$ of outliers. The similarity between the reconstructed images of the fake outliers to the generated samples indicates the ability of a model to project an outlier to a manifold learned from normal samples. Figure~\ref{fig:comparison_beta_fakeoutliers} shows that the reconstructed images and the sampled images look different when $\beta = 0.0003$. When $\beta = 0.0009$, both sets of images look the same but the model lacks variability and yield the same result regardless of the input. On the other hand, the reconstructed images and the generated samples start looking similar when $\beta = 0.0005$ and they become indistinguishable when $\beta = 0.0007$ while keeping the variability. In fact, the maximum value of the heatmap in Figure~\ref{fig:heatmap_roc} (b) is achieved at $\beta=0.0007$ when the percentage of outliers is $10 \%$.

\subsection{Experiment 3: Detecting abnormality in brain images using RVAE}
Recently machine learning methods have been introduced to accelerate the identification of abnormal structures in medical images of the brain and other organs \cite{baur2018deep}. Since supervised methods require a large amount of annotated data, unsupervised methods have attracted considerable attention for lesion detection in brain images. A popular approach among these methods leverages VAE for anomaly detection \cite{chen2018unsupervised} by training the VAE using nominally normal (anomaly free) data. However, if outliers, lesions or dropouts are present in the training data, VAEs cannot distinguish between normal brain images and those with outliers. Here, we tackle this real-world problem by investigating the effectiveness of the RVAE for automated detection of outliers using both simulated and real outliers. We used the VAE architecture proposed in \cite{larsen2015autoencoding} and 20 central axial slices of brain MRI datasets from 119 subjects from the Maryland MagNeTs study of neurotrauma (https://fitbir.nih.gov). We split this dataset into 107 subjects for training and 12 subjects for testing. The experiments using simulated outliers consist of $10 \%$ of two types of outliers: random data dropout (lower intensity lines with a thickness of 5 pixels), and randomly generated simulated lesions (higher intensity Gaussian blobs). For experiments with real outliers, we used 142 central axial slices of 24 subjects from the ISLES (The Ischemic Stroke Lesion Segmentation) database \cite{maier2017isles}. We used 21 subjects for training and 3 subjects for testing. In experiments both with simulated and real outliers, unlike VAE, our RVAE is robust to the outliers in the sense that they are not reconstructed in the decoded images and can, therefore, be detected by differencing from the original images (Figure~\ref{fig:brainlesions}). Due to the small number of samples and more variability and noise in the data, the quality of the reconstructions of examples from the dataset with real outliers is worse than that for dataset with simulated outliers. For quantitative comparison, we apply a pixel-wise ROC study for the data contaminated with simulated outliers. We use reconstruction error in the FLAIR images to generate the ROC curves. The area under the ROC curve was 0.20 for VAE and 0.85 for RVAE which quantitatively demonstrates the success of our RVAE compared to the VAE for lesion detection.
 
\section{Conclusion and Discussion}
The presence of outliers in the form of noise, mislabeled data, and anomalies can impact the performance of machine learning models for labeling and anomaly detection tasks. In this work, we developed an effective approach for learning representations, RVAE, to ensure the robustness of models to the outliers. Our approach relies on the notion of $\beta$-divergence from robust statistics. We formulated cost functions both for Bernoulli and Gaussian cases. Furthermore, we provided an approach to select the robustness hyper-parameter $\beta$ in RVAE based on visual inspection of the generated samples. In the future, this process could be automated by borrowing ideas from adversarial training \cite{kurakin2016adversarial}.
We demonstrated the effectiveness of our approach using both benchmark datasets from computer vision and real-world brain imaging datasets. Our experimental results indicate that the RVAE is robust to outliers in representation learning and can also be useful for outlier detection. Our approach can be applied to the problem of automated anomaly detection in medical images. As future work, our formulation can be extended beyond VAEs to the GANs  framework \cite{goodfellow2014generative} by optimizing a divergence robust to outliers.
In addition to the $\beta$-divergence approach described here we also provide an equivalent formulation for gamma divergence in the supplemental material. 
\bibliographystyle{unsrt}

\bibliography{ref}

\end{document}